%% file: main_arxiv.tex
\documentclass{article}

\usepackage{arxiv}
\usepackage[utf8]{inputenc} 
\usepackage[T1]{fontenc}    
\usepackage{url}            
\usepackage{booktabs}       
\usepackage{amsfonts}       
\usepackage{nicefrac}       
\usepackage{microtype}      
\usepackage{graphicx}
\usepackage{natbib}

\AtBeginDocument{%
  }

\usepackage{subfigure}
\usepackage{booktabs} 
\usepackage{algorithm, algorithmic}
\usepackage[export]{adjustbox}

\usepackage[colorlinks,linkcolor=red,anchorcolor=blue,citecolor=blue]{hyperref}

\usepackage{amsmath}
\usepackage{amssymb}
\usepackage{mathtools}
\usepackage{amsthm}

\theoremstyle{plain}

\theoremstyle{definition}

\theoremstyle{remark}

\input{_packages}

\input{_notation}

\begin{document}


\title{GEANN: Scalable Graph Augmentations for \\ Multi-Horizon Time Series Forecasting}



\author{Sitan Yang
    \thanks{Amazon Forecasting Science. Correspondence to: sitanyan@amazon.com} 
    \And
Malcolm Wolff$\phantom{a}^*$
\And
Shankar Ramasubramanian$\phantom{a}^*$
\And
Vincent Quenneville-Belair$\phantom{a}^*$
\AND
Ronak Metha
\thanks{University of Washington Department of Statistics. Seattle, WA.}
\And
Michael W. Mahoney$\phantom{a}^*$
}

\maketitle

\begin{abstract}
    \input{abstract}
\end{abstract}


\section{Introduction}
\label{sec:introduction}
\input{intro}


\section{Methods}\label{sec:methods}
\input{methods}

\section{Results}\label{sec:results}
\input{experiments}


\section{Conclusion}\label{sec:conclusion}
\input{conclusion}

\bibliographystyle{ACM-Reference-Format}
\bibliography{bibliography}

\appendix

\input{appendix}
\end{document}

%% file: _packages.tex
\usepackage{amsmath,amsthm,amssymb, bbm}
\usepackage{graphicx}
\usepackage{float}
\usepackage[mathscr]{euscript}
\usepackage{enumitem}
\usepackage{bm}
\usepackage{float}
\usepackage{algorithmic}

\usepackage{charter}
\usepackage{setspace}
\usepackage{hyperref}
\hypersetup{
    colorlinks=true,
    linkcolor=blue,
    filecolor=magenta,      
    urlcolor=cyan,
    }

%% file: _notation.tex
\newcommand{\R}{\mathbb{R}}

\theoremstyle{plain}

\theoremstyle{definition}

\theoremstyle{remark}

\newcommand{\p}[1]{\ensuremath{\left(#1\right)}}

\newcommand{\mc}[1]{\ensuremath{\mathcal{#1}}}

\newcommand{\pow}[1]{\ensuremath{^{(#1)}}}

\DeclareMathOperator*{\encoder}{Encoder}
\DeclareMathOperator*{\decoder}{Decoder}
\DeclareMathOperator*{\loss}{Loss}
\DeclareMathOperator*{\gnn}{GNN}
\DeclareMathOperator*{\gem}{GEM}

\DeclareMathOperator*{\corr}{Corr}

\newcommand{\x}{\ensuremath{\mathbf{x}}}
\newcommand{\y}{\ensuremath{\mathbf{y}}}

\newcommand{\denc}{\ensuremath{d_{\text{Enc}}}}
\newcommand{\dgnn}{\ensuremath{d_{\text{GNN}}}}

\newcommand{\X}{\ensuremath{\mathbf{X}}}
\renewcommand{\H}{\ensuremath{\mathbf{H}}}

\newcommand{\mini}{\ensuremath{\mc{M}}}
\newcommand{\expmini}{\ensuremath{\overline{\mc{M}}}}
\DeclareMathOperator*{\hopsubgraph}{GetHopSubgraph}

\renewcommand{\c}{\ensuremath{\mathbf{c}}}

%% file: abstract.tex
Encoder-decoder deep neural networks have been increasingly studied for multi-horizon time series forecasting, especially in real-world applications. 
However, to forecast accurately, these sophisticated models typically rely on a large number of time series examples with substantial history. 
A rapidly growing topic of interest is forecasting time series which lack sufficient historical data---often referred to as the ``cold start'' problem. In this paper, we introduce a novel yet simple method to address this problem by leveraging graph neural networks (GNNs) as a data augmentation for enhancing the encoder used by such forecasters. 
These GNN-based features can capture complex inter-series relationships, and their generation process can be optimized end-to-end with the forecasting task. 
We show that our architecture can use either data-driven or domain knowledge-defined graphs, scaling to incorporate information from multiple very large graphs with millions of nodes. 
In our target application of demand forecasting for a large e-commerce retailer, we demonstrate on both a small dataset of 100K products and a large dataset with over 2 million products that our method improves overall performance over competitive baseline models. More importantly, we show that it brings substantially more gains to ``cold start'' products such as those newly launched or recently out-of-stock.

%% file: intro.tex
Recent years have shown a growing interest in using deep neural networks (DNNs) for multi-horizon time series forecasting problems \citep{Lim2018, Madeka2018,  li2018diffusion, Wen2019,  Benidis2022Deep}.
This is due to their flexibility of consuming various types of inputs and to their large model capacity to effectively learn on a huge amount of data, compared to traditional methods. Canonical DNNs (e.g., LSTM \citep{hochreiter1997long} and GRU \citep{cho2014learning}) have achieved success in many impactful real-world applications \cite{Carlos2016, Bose2017, Zhang2018MultistepPF}, and most recent methods in this field leverage Convolutional Neural Networks \citep[e.g.,][]{Wen2017AMulti,borovykh2017conditional,sezer2018algorithmic,chen2020probabilistic} and Transformer architectures \citep[e.g.,][]{Lim2018,li2019enhancing,wu2020deep,Eisenach2020MQTransformer,lim2021temporal} to further improve performance.

While a primary benefit of forecasting with DNNs is the ability to train on a large number of time series with long histories, these models also tend to rely on that scale to effectively forecast.
Difficulties can arise, however, when there is a significant lack of historical data. This is often referred to as the ``cold start'' problem, and it has garnered recent interest in time series forecasting literature \citep[e.g.,][]{moon2020solving, aguilar2020cold, bottieau2022cross, chauhan2020time}. As many modern DNN forecasting models adopt a Seq2Seq structure \cite{Sutskever2014}, which assumes that the observed time series are uncorrelated, they tend to produce poor forecasts for ``cold start'' series. In these cases, learning relational information across series may bring additional performance gains to forecasting models.

Recent research has explored the use of Graph Neural Networks (GNNs) to capture complex inter-series relationships.  While graph data augmentation has been studied extensively for general graph-based tasks such as node classification \cite{liu2021, ParkNIPS2021} and edge prediction \cite{WangNIPS2021}, its use in time series forecasting remains relatively under-explored. GNNs for time series explicitly model the relationship between observations by representing time series as nodes and their interactions as edges in a graph, and this has shown promising results in early work \cite{yu2017spatio,li2017diffusion}. More recently, \citep{YangKDD2022} introduced using a retrieval mechanism to augment time series, achieving significant performance gains for multi-horizon forecasting. However, most major work in this field \cite[see][]{Zugner2021AStudy} have only used a single graph with no more than 1,000 nodes in evaluation, proposing methods which fail to scale to many practical applications. To alleviate this limitation, a few recent advances have improved the scalability of GNNs for time series forecasting by using altered, smaller graphs \citep{Gandhi2021Spacio-Temporal, satorras2022multivariate, Huang2021Scaling} and mini-batch sampling algorithms \citep{Hamilton2017Inductive, Chang2019Cluster-GCN}. 
 
In this paper, we expand on this literature by proposing a novel methodology---GEANN (``Graph Ensemble Augmented Neural Networks'')---as a practical DNN solution to the ``cold start'' problem in large-scale time series forecasting. GEANN is a parsimonious model using GNNs as a data augmentation mechanism to enhance encoders typically used in the Seq2Seq forecasting architectures. We apply our method to the sequence structure of MQ-CNN from the family of MQ-Forecaster models \cite{Wen2017AMulti, Eisenach2020MQTransformer}, which have shown state-of-the-art performances for time series, and especially for our target application -- demand forecasting. We leverage the forecasting quality of MQ-CNN with graph-encoded information as an add-on component to improve representation learning. GEANN can scale to one or more very large graphs, which we demonstrate can lead to substantial performance improvements. Graphs in GEANN are predefined and static, and we can optionally assemble them using any domain-specific knowledge. In addition, we propose the use of pre-computed sparse graphs and their induced subgraphs to parallelize the GNN learning process on large and complex datasets.

To our best knowledge, this work is the first to show the benefit of GNNs as a data augmentation for time series forecasting in large scale real-world applications. We evaluate the proposed method in our target application --  demand forecasting for a large e-commerce retailer, using both a small dataset consisting of $\sim$100K products and a large application with over 2MM products. In both cases, we observe overall performance improvements over competitive MQ-Forecaster baselines. More importantly, we demonstrate that our method brings substantially larger gains in ``cold-start'' scenarios, such as new products with little to no sales history and recently out-of-stock products with sales history being incorrectly suppressed due to inventory constraints.

%% file: methods.tex
\subsection{Problem Formulation} 
Here and for the remainder of the paper, we denote tensors in boldface, matrices in upper case, and vectors in lower case. Let $Y\in\mathbb{R}^{N\times T}$ denote $N$ time series of length $T$ as targets, $\mathbf{X}^{(t)} \in\mathbb{R}^{N\times T \times d}$ a set of $d$ time series covariates, and $X^{(s)} \in \mathbb{R}^{N\times m}$ a set of $m$ static covariates. Given a \emph{context length} $C \geq 0$---i.e. the number of past observations used for modeling from the forecast time $t$---and a collection of \emph{horizons} $\mathcal{H}$ to forecast in the future, we wish to generate the conditional forecast given $\X_{t-C:t}^{(t)} = (X_{t-C}^{(t)},..., X_{t}^{(t)})$, $Y_{t-C:t}$, and $X^{(s)}$ via the model
\begin{equation}
\begin{aligned}
	\widehat{Y}_{t, \mathcal{H}} = f\left(Y_{t-C:t}, \mathbf{X}_{t-C:t}^{(t)}, X^{(s)} ; \bm{\theta}\right),
	\label{eqn:dependence_aware}
\end{aligned}
\end{equation}
where $\bm{\theta}$ represents a collection of learnable parameters. To promote scalability, current deep learning architectures often consume the information of each observation $i$ independently with the shared set of parameters $\bm{\theta}$ as
\begin{equation}
\begin{aligned}
\widehat{y}_{i,t,\mathcal{H}} = f\left(y_{i,t-C:t}, X_{i,t-C:t}^{(t)}, x_i^{(s)}\ ;\ \bm{\theta}\right).
\label{eqn:indep_aware}
\end{aligned}
\end{equation}
This treatment lends itself naturally to parallel computing, but it discards relational dependencies which may exist across time series, motivating the use of GNNs for improving equation \eqref{eqn:indep_aware} to learn such information. 

GNN layers used for time series in this paper are generally of the form
\begin{align*}
	g_{i, t} &= {\gnn}_{\theta}\p{H_{t}\ ;\ \mathcal{G}},
\end{align*}
where the inputs $H_t \equiv (h_{i, t})_{i=1}^N \in \mathbb{R}^{N\times \denc}$ are intermediate embeddings, $\mathcal{G}$ is a graph describing pairwise relationships among observations, and the output is a graph-aware embedding. However, current GNN methods typically operate by modeling pairwise relations among observations across the entire graph simultaneously, leading to poor scalability for large graphs.

After the model is chosen, the parameters are tuned to optimize the loss during training as
\begin{align}
	\loss(\bm{\theta}) = \sum_i \sum_h \sum_t \ell(y_{i, t, h}, \widehat{y}_{i, t,h}) .
	\label{eqn:loss}
\end{align}
The quantile loss (QL) function used in our experiments is detailed in Appendix \ref{app:dqf}. 
\begin{figure*}
	\centering
	\includegraphics[width=0.85\linewidth]{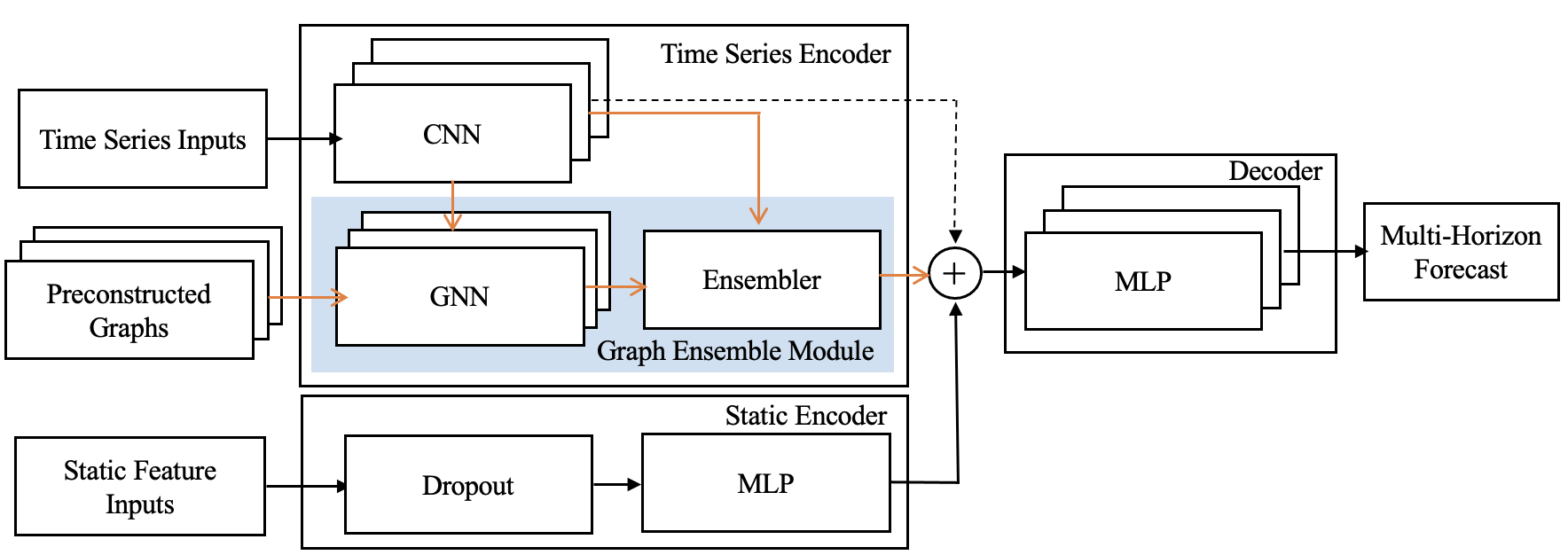}
	\caption{Based on the encoder-decoder structure of MQ-CNN \cite{Wen2017AMulti}, the embedded graph ensemble module of GEANN maps the time series encoded input $h_{i, t}$ through a number of GNN operations (details below). The outputs are then used to form the ensemble representation $g_{i, t}$, which are decoded together with $h_{i, t}$ into forecasts.}
	\label{fig:architecture}
\end{figure*}

\subsection{Model Architecture} 
Figure \ref{fig:architecture} summarizes our architecture. GEANN adopts the encoder-decoder architecture of MQ-CNN (as detailed in \cite{Wen2017AMulti}) with our \emph{graph ensemble module} (GEM) embedded. The encoder uses a stack of dilated temporal convolutions to summarize past targets and time-varying covariates into a sequence of hidden states $H_t \equiv (h_{i, t})_{i=1}^N \in \mathbb{R}^{N\times \denc}$. The proposed GEM component encodes $H_t$ to additional hidden states $G_t \equiv (g_{i, t})_{i=1}^N \in \R^{N\times \dgnn}$,
\begin{align*}
    \text{GEM}_{\bm{\theta}}(H_t ; \mathcal{G}^{(1)},\ldots,\mathcal{G}^{(R)}).
\end{align*}
through $R$ individual GNN operations, each of which on separate fixed graphs $\mathcal{G}^{(r)}$. That is, for $r=1,...,R$,
\begin{align*}
	g_{i, t}^{(r)} &= {\gnn}^{(r)}_{\theta}\p{H_{t}\ ;\ \mathcal{G}^{(r)}}.
\end{align*}
The encoded states from all GNN layers are combined with trainable weights $w^{(r)} \geq 0, \sum_{r=1}^R w^{(r)} = 1$ to yield the final output:
\begin{align*}
    	g_{i, t} &= \sum_{r=1}^R w^{(r)} g^{(r)}_{i, t} \quad \forall i.
\end{align*}
Each graph remains static, but the node embeddings are dynamic across all time steps $t$. 

Many prior methods and associated GNN libraries for graph learning such as DGL \citep{wang2019dgl} and PyTorch Geometric \citep{Fey2019} typically require the entire node embedding set $H_t$ and the whole graph structure $\mathcal{G}^{(r)}$ during each forward pass, regardless of batch size, which is infeasible for large datasets. We instead propose a learning algorithm for GEANN (see details in Appendix \ref{app:learningalgo}) which applies ``top-k'' neighborhood sampling and induces ``$L$-hop'' subgraphs; this uses nodes in each mini-batch as seed nodes to alleviate the scaling issue (see, e.g. \cite{Hamilton2017Inductive,Chang2019Cluster-GCN} for related techniques). We note that the graph sparsity controls the use of GPU memory during training.

The resulting embeddings $G_t$ are then combined with $H_t$ for augmenting the representation learning process. Notice that this process is done for each $t$ separately during training, and the parameters are optimized together with the subsequent forecasting task. 

In this paper, we adopt the same decoder as in MQ-CNN, but we note that any other decoder can be used. Formally, our GEANN architecture is as follows:
\begin{align*}
	& h_{i, t} = {\encoder}_{\bm{\theta}}^{(t)}\p{y_{i, t-C:t}, X_{i, t-C:t}^{(t)}},\\
	& h_{i}^{(s)} = {\encoder}_{\bm{\theta}}^{(s)}\p{x_i^{(s)}},\\
	&G_{t} = {\gem}_{\bm{\theta}}\p{H_{t} ; \mathcal{G}^{(1)}, ..., \mathcal{G}^{(R)}}, \\
	&\hat{y}_{i,t,\mathcal{H}} = {\decoder}_{\bm{\theta}}(h_{i, t}, g_{i, t}, h_{i}^{(s)}).
\end{align*}
For the GNN operation, we use $L$ graph convolutional network (GCN) layers  \citep{Henaff2015DeepConvolutional} for a $L-$hop neighborhood configuration, but this can be easily extended to other types of graph learning layers such as graphSAGE \cite{Hamilton2017Inductive} or GAT \cite{vel2018graph}. 

\paragraph*{\bf Graph Construction} 
GEANN can use any predefined graph along with optionally a specified weight for each edge. Here we model each of $N$ time series in $H_t, t=1,\ldots,T$ as a node and their interactions as edges, and we generally consider two types of graphs: the data-driven graph, and the domain knowledge-defined graph. GEANN natively supports sparse graphs which can be efficiently stored as edge lists. For each graph constructed, we further apply an additional ``top-k'' operation during training to control the GPU memory usage of each mini-batch so that the process can be efficiently parallelized.

We construct a data-driven graph using a similarity metric between node $i$ and $j$. In this paper, we choose the Pearson correlation coefficient as the metric, but other metrics can also be used. We adopt the same method used in \cite{YangKDD2022} to base our calculations on the embedding vectors generated by a pretrained MQ-CNN model as 
$ S_{\text{Corr}}(i, j) = \left|{\corr}\p{H^{(0)}_{i}, H^{(0)}_{j}}\right|.$
Here $H^{(0)}_{i}$ denotes the pretrained version of $h_{i, t}, t=1,...,T$ from a frozen MQ-CNN model, and $\text{Corr}(\cdot,\cdot)$ indicates the Pearson correlation coefficient. Based on the similarity metric we further obtain a $k$-NN embedding graph.

We also consider constructing graphs with domain-related knowledge available. For online marketplace demand forecasting, there often exists a rich set of relational information (e.g., site catalog information and customer browsing data). In this paper, we use \textit{browse nodes}, which are attributes visible on the website of many e-commerce retailers to help customers navigate through the vast selection of products. For example, browse node set for a men's fashion sweatshirt consists of ``Clothing, Shoes \& Jewelry'', ``Men'', ``Clothing'' and ``Fashion Hoodies \& Sweatshirts.''  These correspond to the nodes visited in the browsing tree to locate the product. Thus, products that belong to the same browse node are likely to be co-browsed by customers, providing a notion of both substitutable and complimentary goods. 

We note that unlike the ``ground-truth'' graph used in spatio-temporal tasks such as the traffic flow forecasting \cite{li2017diffusion, yu2017spatio}, the graph derived from browse nodes or other similar attributes only indicates one of many complex and subtle relationships existing among data, and it is inherently more difficult to show the value-add of graph learning in this case.

%% file: experiments.tex
In this section we evaluate GEANN on two demand datasets from a large e-commerce retailer, that include time series features such as unit sales, promotions, holidays and detail page views as well as static metadata features such as catalog information. Similar datasets with the same set of features but generated in different time windows have been used in \cite{Wen2017AMulti, Eisenach2020MQTransformer, YangKDD2022}. We have obtained five years (2016-2021) of history for time series data. Each model is trained on three years (2016-2019) of demand data, one year is held out for validation, and the final year is kept for evaluation. The task is to forecast the 50th and 90th quantile of weekly demand for up to one year at each of the 52 forecast creation time (see Appendix \ref{app:dqf} for metric~details).

GEANN is implemented using the Pytorch framework with 8 NVIDIA V100 Tensor Core GPUs. The model is optimized using AdamW \citep{Kingma2014} with default parameters. We limit our node neighborhood to a depth of 2 with a maximum neighborhood size of 10. For the graph encoding, we use a 2-layer GCN model with 32 hidden units.

\subsection{Small Scale Experiment}
The first dataset consists of 100K products with the largest number of total units sold during the training period from 5 main European marketplaces (EU5), and we compare the following architectures:
\begin{itemize}
    \setlength\itemsep{-.1em}
	\item \textbf{MQ-CNN}: the MQ-CNN model \cite{Wen2017AMulti}
	\item \textbf{MQ-T}: the MQ-Transformer model \cite{Eisenach2020MQTransformer}
	\item \textbf{GEANN-bw}: GEANN using a browse node graph
	\item \textbf{GEANN-kNN}: GEANN using a kNN embedding graph
	\item \textbf{GEANN-bw+kNN}: GEANN using both a browse node and kNN embedding graph
\end{itemize}
MQ-CNN and MQ-T serve as the baseline models in the comparison.\footnote{We mainly focus our comparison on MQ-CNN, as MQ-T requires substantially more GPU memory, and we are already memory bound for GEANN. Moreover, we see that the improvement of MQ-T on MQ-CNN is orthogonal to those in GEANN, and thus they can be combined in the future study.} For GEANN, each graph contains 100K nodes. For the browse node graph, we rank all other products related to a particular product by the number of times where they appear together in the same browse node, and we choose the top 10 products as the neighbors. For the kNN embedding graph, we consider the 10 nearest neighbors calculated on the pretrained embeddings of MQ-CNN. We also include the performance of GEANN with both graphs. Furthermore, for the purpose of ablation, we include 3 additional model configurations: 
\begin{itemize}
	\item MQ-CNN-L: MQ-CNN with the number of parameters matching that of GEANN by increasing the dilation capacity of the CNN layer
	\item GEANN-idm: GEANN with a ``zero-neighbor'' graph (i.e., adjacency matrix being an identity matrix) that contains no additional graph-related information 
	\item GEANN-random: GEANN with a randomly connected graph with no predictive information expected
\end{itemize}

We train each model to 100 epochs using batch size of 256, and each model evaluation is computed with 5 identical runs using different random seeds. We summarize each model performance in Table \ref{tab:small-exp-res}.
\begin{table}[!ht]
\caption{Performance metrics on 100K EU5 retail products. The results are rescaled so that they are relative improvements over MQ-CNN. Lower is better. The number of parameters used by each model is also included.}
	\label{tab:small-exp-res}
	\begin{center}
		\begin{tabular}{lllll}
			\toprule
			Model        & P50 QL   & P90 QL  & Overall & Param  \\
			\midrule
			MQ-CNN        & 1.000 & 1.000 & 1.000 & 850k \\
			MQ-T      	& 1.060  & 0.999 & 1.029& 858k\\
			\midrule
			MQ-CNN-L &  1.059 & 1.047 & 1.053 & 898k\\
			GEANN-idm &1.005  & 0.1003 & 1.004& 900k \\
			GEANN-random & 1.003 & 0.993 &  0.998& 900k\\
			\midrule
			GEANN-bw 	& 0.977 & {\bf 0.974} & {\bf 0.976}& 900k\\
			GEANN-kNN   & 0.979 & 0.977 & 0.978& 900k\\
			GEANN-bw+kNN & {\bf 0.969} & 0.983 & {\bf 0.976}& 915k\\
			\bottomrule
		\end{tabular}
	\end{center}
\end{table}
We observe all GEANN variants lead to noticeable overall performance improvements ($\sim$2\%-3\%) over baselines. Notably, the three ablations do not seem to improve upon baselines, even with similar number of parameters compared to GEANN. Hence, the predictive information in the browse node and $k$-NN embedding graph cannot be produced from a randomly generated graph. GEANN-idm is expected to be similar to MQ-CNN, and indeed it has on-par performance. The ensemble of the two graphs does not seem to produce further accuracy gain in our experiment.
\subsection{Large Scale Application}
We test on a large scale application of demand forecasting that involves over 2MM products with most sales from North America marketplaces (NA). Leveraging a graph with millions of nodes is a rare challenge considered in the previous graph-based methods \cite{wu2020connecting, shang2021}. Again using MQ-CNN and MQ-T as baselines, we compare the test performance of GEANN-bw, GEANN-kNN and GEANN-bw+kNN.
We train each model with a batch size of 512. Each epoch takes around 30 minutes for the model with a single graph and 1 hour with two graphs. Each model evaluation is averaged across 3 identical runs, and test results are summarized in Table \ref{tab:na-retail-exp-res}.
\begin{table}[!ht]
	\caption{Performance metrics on 2 million NA retail products. The results are rescaled so that they are relative improvements over MQ-CNN. Lower is better.}
	\label{tab:na-retail-exp-res}
	\begin{center}
		\begin{tabular}{llll}
			\toprule
			Model        & P50 QL  & P90 QL & overall    \\
			\midrule
			MQ-CNN        & 1.000 & 1.000 & 1.000 \\
			MQ-T      	&  0.999 & 0.998 & 0.998 \\
			\midrule
			GEANN-bw 	& 0.997 & {\bf 0.985} & {\bf 0.991}  \\
			GEANN-kNN   & 1.023 & 1.044 & 1.033 \\
			GEANN-bw+kNN & {\bf 0.993} & 1.001 & 0.997\\
			\bottomrule
		\end{tabular}
	\end{center}
\end{table}
We notice that GEANN improves over baselines with the browse node graph, but in this case it degrades with the kNN graph, which is likely to confirm the predictive information from browsing data. A fundamental question related to degraded performance when using the data-driven graph is whether it is a function of the model architecture or graph estimation procedure. Further analysis of the data-driven graph suggests the data-driven graph is highly volatile when $k$ is small, providing a low signal-to-noise ratio (detailed in Section \ref{sec:graph_analysis}). The two graph ensemble method performs on-par with the baselines mainly due to the under-performance of the kNN~graph. 

\paragraph*{\bf Newly launched and Recently Out-Of-Stock Products} 
For GEANN-bw, in addition to overall performance gain, we also find that it significantly improves for two challenging and important groups: newly launched, and recently out-of-stock (OOS) products. Forecasting demand for new products is generally difficult due to little to no time series history. For recently OOS products, past sale history during OOS period is no longer a good signal for demand prediction due to inventory constraint. For these two cases, we show in Table \ref{tab:na-retail-oos} that GEANN brings substantially more gains ($\sim$5\%) compared to the overall improvement. Our empirical analysis indicates that these accuracy improvements tend to stem from faster calibration of demand forecasts for new and OOS products, by adjusting their forecasts based on the historic demand of similar products used as neighbors in GEANN. Hence by incorporating relational information, our model is able to leverage greater contextual understanding to quickly calibrate forecasts for cold-start products.
\begin{table}[!ht]
	\caption{Model performance comparison for newly launched and recently out-of-stock products relative to MQ-CNN.}
	\label{tab:na-retail-oos}
	\begin{center}
		\begin{tabular}{lccc}
			\toprule
			 Model  & \multicolumn{3}{c}{GEANN-bw} \\
			 \cmidrule(lr){2-4}
			         & P50 QL & P90 QL & overall \\
			\midrule
			Newly Launched  &  0.975 &  0.910 & 0.943  \\
			Recently OOS &   0.974 & 0.957 & 0.965 \\
			\bottomrule
		\end{tabular}
	\end{center}
\end{table}

\subsection{Data-Driven Graph Stability}
\label{sec:graph_analysis}
In this paper, we have considered both data-driven and domain-knowledge defined graphs.
We observe that the performance of using a data-driven graph is volatile, while the domain-knowledge graph seems to produce consistent improvements across different datasets. In this section, we conduct an in-depth analysis for the data-drive graph used to provide insights for this result.
\begin{figure}[H]
    \centering
    \hspace{-.2in}\includegraphics[scale=.45]{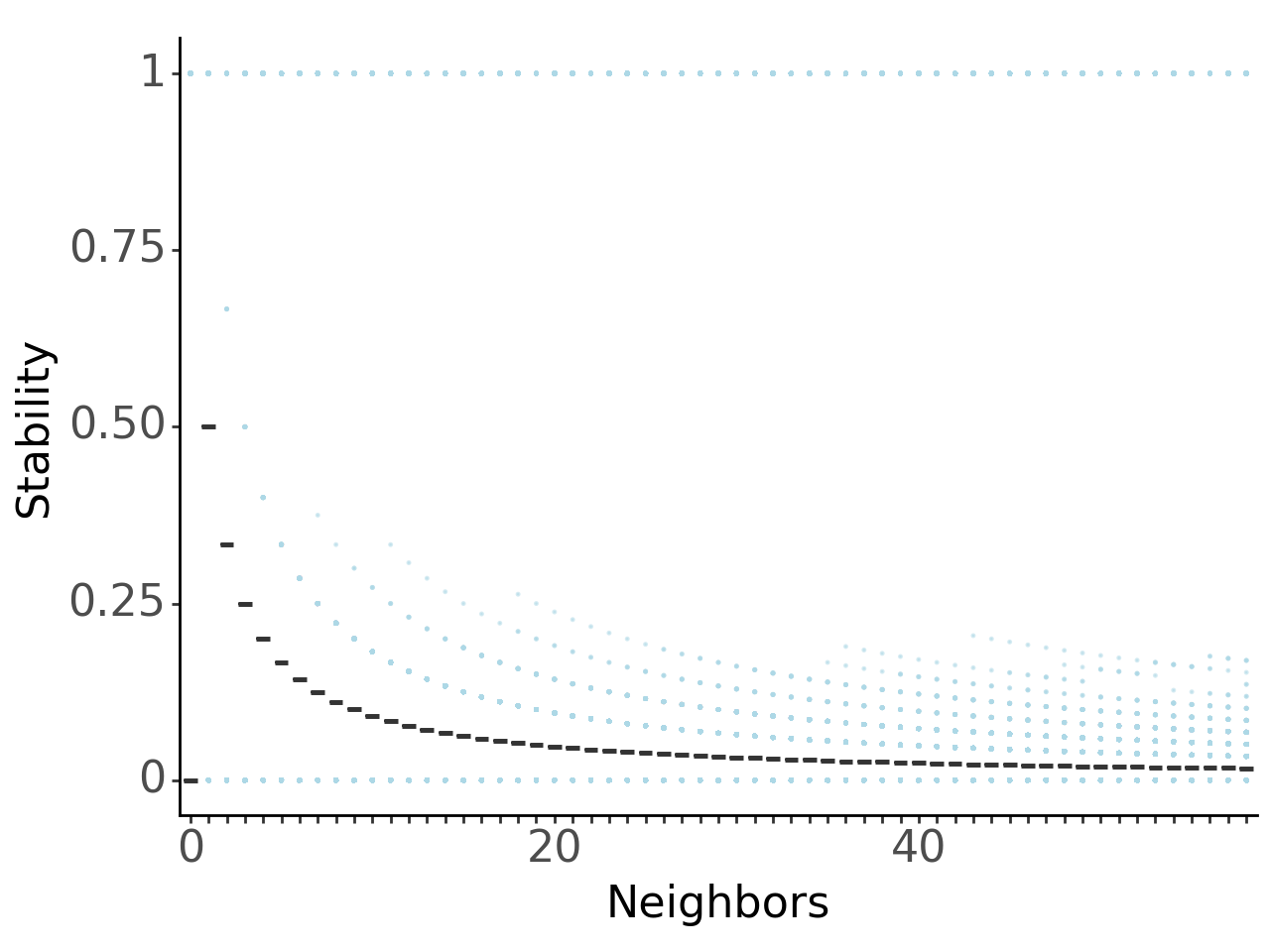}
    \caption{Boxplots of stability by number of neighbors for $k$-NN generated from Pearson correlation between embeddings $\mathbf{H}$ across 3 runs of MQ-CNN. Blue points represent~outliers.}
    \label{appendix:stability}
\end{figure}
\paragraph*{\bf Distribution of similarity metrics} 
We first compare the nearest neighbor distribution for these two graphs, respectively, for the large scale dataset used in Section 3.2. Appendix Figure \ref{fig:knn-stats} depicts a histogram of the estimated means and standard deviations of the Pearson correlations from 10 nearest neighbors for each product in GEANN-kNN. The substantial concentration of mean correlations near unity, and similarly the concentration of standard deviation near zero, show that many of the products have nearly interchangeable rank. On the other hand, Appendix Figure \ref{fig:bw-stats}, which shows a similar histogram as those for co-browsing counts from 10 nearest neighbors for each product in GEANN-bw, presents much stronger variation. Moreover, it is clear that a subset of products have strong similarity relative to others, with mean co-browsing counts above~35.
\paragraph*{\bf On $k$-NN graph volatility} 
One reason such context-based $k$-NN graphs may show unreliable relationships relevant to forecasts is the stability of the $k$-NN construction on embeddings $\mathbf{H}$. We define stability for a product $a$ and number of neighbors $k$ across runs $r$, each run generating a context $\mathbf{H}^{(r)}$,~as
\begin{equation}
    \nonumber
    \begin{aligned}
    \text{Stability}(a ; k, \mathcal{R}) &\equiv \dfrac{\left|\bigcap_{r\in\mathcal{R}} \text{KNN}_{r}(a ; k)\right|}{k}.
    \end{aligned}
\end{equation}
Stability is a measure of preservation of graph similarity across model training runs; a random selection of neighbors has a stability proportional to a HyperGeometric$(N, k, k)$, where $N$ is the number of data observations. When $N$ is reasonably large, the stability is close to 0, and a deterministic graph has a stability of 1.0. Prior work in natural language processing field shows the stability of frequently used word embeddings are often 0.8 or greater \citep{borah2021word}.

Figure \ref{appendix:stability} shows the distribution of stability by the number of neighbors for our embedding based $k$-NN using Pearson correlation across 3 model runs. We find generally low stability of $k$-NN relationships generated from the embeddings $\mathcal{H}$. While the stability of low numbers of neighbors is relatively larger at 0.5, it quickly decreases to represent an approximately random graph as the number of neighbors increases--- likely due to orderings based on correlation being overcome by model noise. Moreover, the stability is highly consistent across products. A maximal stability of approximately 0.5 suggests that the $k$-NN based graph may provide an inconsistent signal due to noise in the embeddings $\mathbf{H}$.

%% file: conclusion.tex
In this paper, we demonstrate that for demand forecasting our proposed graph-based time series forecasting method GEANN outperforms the current Seq2Seq models while maintaining the scaling advantage. One interesting future direction is to consider different GNN architecture such as the graph attention network architecture~\cite{Velickovic2017Graph}. In addition, more sophisticated graph construction methods can be used, such as choosing graphs and model parameters based on their Network Community Profile curves~\cite{Faerman2017LASAGNE}.

%% file: appendix.tex
\section{Distributional Quantile Forecasts}
\label{app:dqf}
We refer to distributional forecasts as quantile forecasts, and we consider our target application (demand forecasting) as predicting certain quantiles of the future demand distribution at a weekly grain for up to one year in the future. 
A natural metric measuring how accurate a quantile forecast $f$ at the $q$-th quantile with respect to the true demand $d$ is the quantile loss $L$ (QL), as defined below:
\begin{align}
\label{eq:loss_1}
L_q(d, f) = q(d-f)_+ + (1-q)(f-d)_+,
\end{align}
where $(\cdot)_+= \max(\cdot,0)$.
This metric is typically aggregated across samples and time horizons, and it is weighted by their actual demand. The weighted P50 and P90 QL are used to measure the quality of distributional forecasts as the 50th and 90th percentile of the demand distribution. The same metric has been used for previous work on multi-horizon time series forecasting \cite{Wen2017AMulti, Lim2018, Eisenach2020MQTransformer, YangKDD2022}.

\section{GEANN Learning Algorithm}
\label{app:learningalgo}
In this section, we describe the learning algorithm of GEANN. See the details in Algorithm \ref{alg:geann}. While stochastic gradient descent (SGD) is the optimization algorithm shown, in practice other methods such as Adam \citep{Kingma2014} is often used.
\begin{algorithm}
	\caption{GEANN Learning Algorithm}
	\label{alg:geann}
  \begin{algorithmic}
			\STATE \textbf{Require:} Number of epochs $E$, mini-batch size $m$, number of GNN layers $L$, candidate graphs $\mathcal{G}\pow{1}, \ldots, \mathcal{G}\pow{R}$, context length $C$, maximum horizon $H$, training data $\p{\y_{i, 1:T}, \x_{i, 1:T}, \c_i}_{i \in [N]}$, loss function $\ell$, learning rate $\eta$, initial parameters $\bm{\theta}$.\\
			\FOR{$i_{\text{ep}} \in \{1, \ldots, E\}$}
			\STATE Partition $[N]$ into mini-batches $\mc{M}_1, \ldots, \mc{M}_{N / m}$
			\FOR{$b \in \{1, \ldots, N/m\}$}
			\STATE $h_{i, t} = {\encoder}_{\theta}\p{\y_{i, t-C:t}, \mathbf{X}_{i, t-C:t}, X_i^{(s)}}$
			\FOR{$r \in \{1, \ldots, R\}$}
			\STATE $\expmini\pow{r}, \overline{\mathcal{G}}\pow{r} = \hopsubgraph(\mini_b, \mathcal{G}^{(r)}, L)$.
			\STATE $h\pow{r}_{i, t} = {\encoder}_{\theta}\p{\y_{i, t-C:t}, \X_{i, t-C:t}, X_i^{(s)}}$
			\STATE $\H\pow{r}_{t} = (h_{i, t}^{(r)})_{i \in \expmini\pow{r}}$.
			\STATE $\p{g^{(r)}_{i, t}}_{i \in \expmini\pow{r}} = {\gnn}^{(r)}_{\theta}\p{\H\pow{r}_{t}, \overline{\mathcal{G}}\pow{r}}$.
			\ENDFOR
			\STATE $\mathbf{g}_{i, t} = (g\pow{1}_{i, t},\ldots,g\pow{R}_{i,t})$ for all $i \in \mc{M}_b$.
			\STATE $\p{\hat{y}_{i, t,\mathcal{H}}} = \decoder_{\bm{\theta}}(h_{i,t}, \mathbf{g}_{i, t})$.
			\STATE $\loss(\bm{\theta}) = \sum_{i \in \mini_b}\sum_{t} \ell(y_{i, t, \mathcal{H}}, \hat{y}_{i, t,\mathcal{H}})$.
			\STATE $\bm{\theta} \leftarrow \bm{\theta} - \eta \nabla \loss(\bm{\theta})$.
			\ENDFOR
			\ENDFOR \\					
	\RETURN $\bm{\theta}$.
	\end{algorithmic}
\end{algorithm}
The graph ensemble module first identifies the time series in the current mini-batch (denoted as $\mini_b$). For each graph $\mathcal{G}\pow{r}$, the module determines a subgraph $\overline{\mathcal{G}}\pow{r}$ of $\mathcal{G}\pow{r}$ induced by the ``$L$-hop-out neighborhood'' of the seed nodes $\mini_b$, and containing all nodes $\expmini\pow{r}$ reached by traversing at most $L$ edges on $\mathcal{G}\pow{r}$. This is also the set of nodes reached by applying an $L$-layer graph convolution on seed nodes \cite{Hamilton2017Inductive}.  In this process, $\mathcal{G}\pow{r}$ and $\overline{\mathcal{G}}\pow{r}$ are guaranteed to produce the same output for seed nodes in $\mini_b$, ensuring that a gradient estimated with elements of $\mini_b$ remains unbiased. For large graphs, the neighborhood memberships can be pre-computed and stored offline while only retrieving nodes of $\overline{\mathcal{G}}\pow{r}$ online in each mini-batch, effectively reducing the use of GPU memory. Notice that only the graph representations of seed nodes are used in the decoder and backpropagation step. 
We note that the sparsity of $\mathcal{G}\pow{r}$ is critical for the learning algorithm to be efficiently trained in parallel. This directly controls the size of {\small $\expmini\pow{r}$}. Therefore in this process, we typically apply a ``top-k'' operation on the given graph so that if $\mini_b$ has $m$ nodes, the upper bound for the number of nodes in {\small $\expmini\pow{r}$} is $m(1 + k^L)$.  Consequently we can determine the value of $(m, k)$ based on memory constraints on the GPU. 

\section{Additional Figures for Data-drive Graph Analysis}
\begin{figure}
\centering
	\includegraphics[scale=0.4,valign=t]{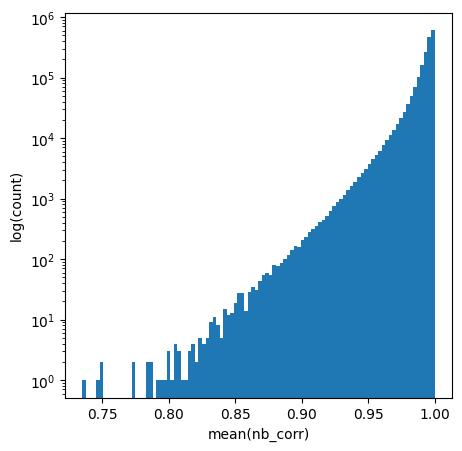}
	\includegraphics[scale=0.4,valign=t]{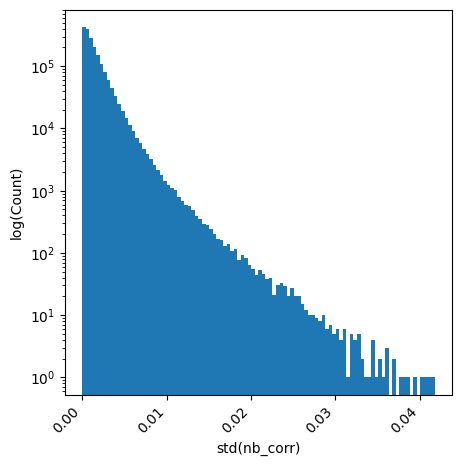}
	\caption{Histogram of the estimated mean and standard deviation of the Pearson correlations in GEANN-kNN.}
	\label{fig:knn-stats}
\end{figure}

\begin{figure}
\centering
	\includegraphics[scale=0.4,valign=t]{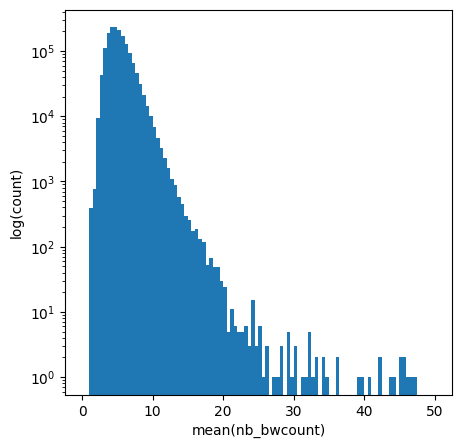}
	\includegraphics[scale=0.4,valign=t]{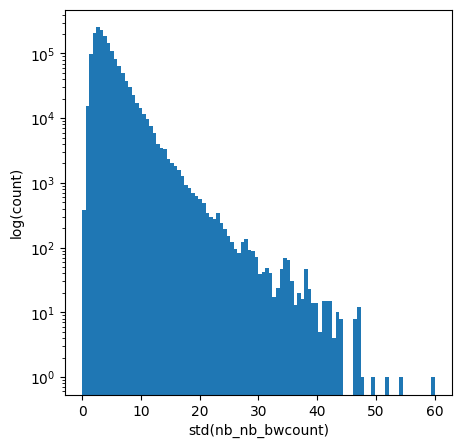}
	\caption{Histogram of the estimated mean and standard deviation of the co-browsing counts in GEANN-bw.}
	\label{fig:bw-stats}
\end{figure}

In this section, we include additional figures for the data-drive graph stability analysis conducted in Section 3.3. 
See Figure~\ref{fig:knn-stats} and Figure~\ref{fig:bw-stats}.
In particular, we calculate using the kNN embedding graph the means and standard deviations of the Pearson correlations from 10 nearest neighbors of each product, and we compare with those calculated from the browse node graph. We note that the score associated in the browse node graph is the co-browsing count, i.e., the number of times a certain product and its neighbor product belong to the same node in the browsing tree, and we use this score to rank all neighbor products for applying the ``top-k'' operation.  

